\documentclass[letterpaper, 10 pt, conference]{ieeeconf}
\IEEEoverridecommandlockouts 
\usepackage{cite}
\usepackage{amsmath,amssymb,amsfonts}
\usepackage{algorithmic}
\usepackage[algo2e]{algorithm2e} 
\usepackage{graphicx}
\usepackage{textcomp}
\let\labelindent\relax
\usepackage[inline]{enumitem}
\usepackage{xcolor}
\usepackage{caption}
\usepackage{graphicx}
\usepackage{algorithm}  

\usepackage{diagbox,slashbox}
\usepackage{subcaption}
\usepackage{hyperref}
\usepackage{verbatim}

\def\BibTeX{{\rm B\kern-.05em{\sc i\kern-.025em b}\kern-.08em
    T\kern-.1667em\lower.7ex\hbox{E}\kern-.125emX}}
    
\newenvironment{myitemize}{\begin{list}{$\bullet$}
{\setlength{\topsep}{1mm}
\setlength{\itemsep}{0.25mm}
\setlength{\parsep}{0.25mm}
\setlength{\itemindent}{0mm}
\setlength{\partopsep}{0mm}
\setlength{\labelwidth}{15mm}
\setlength{\leftmargin}{4mm}}}{\end{list}}

\begin{document}

\title{\LARGE \bf End-to-end Uncertainty-based Mitigation of Adversarial Attacks to Automated Lane Centering\\
}


\author{Ruochen Jiao$^{1,2}$, Hengyi Liang$^{1,2}$, Takami Sato$^{3}$, Junjie Shen$^{3}$, Qi Alfred Chen$^{3}$, Qi Zhu$^{2}$%
\thanks{$^{1}$ These two authors contribute equally to the work.}%
\thanks{$^{2}$ Ruochen Jiao, Hengyi Liang and Qi Zhu are with the Department of Electrical and Computer Engineering, Northwestern University, IL, USA.}%
\thanks{$^{3}$ Takami  Sato, Junjie Shen and Qi Alfred Chen are with the Department of Electrical Engineering and Computer Science, University of California, Irvine, CA, USA.}%
}


\maketitle

\begin{abstract}
In the development of advanced driver-assistance systems (ADAS) and autonomous vehicles, machine learning techniques that are based on deep neural networks (DNNs) have been widely used for vehicle perception. These techniques offer significant improvement on average perception accuracy over traditional methods, however have been shown to be susceptible to adversarial attacks, where small perturbations in the input may cause significant errors in the perception results and lead to system failure. Most prior works addressing such adversarial attacks focus only on the sensing and perception modules. In this work, we propose an end-to-end approach that addresses the impact of adversarial attacks throughout perception, planning, and control modules. In particular, we choose a target ADAS application, the automated lane centering system in OpenPilot, quantify the perception uncertainty under adversarial attacks, and design a robust planning and control module accordingly based on the uncertainty analysis.  
We evaluate our proposed approach using both public dataset and production-grade autonomous driving simulator. The experiment results demonstrate that our approach can effectively mitigate the impact of adversarial attack and 
can achieve $55\%\sim90\%$ improvement over the original OpenPilot.
\end{abstract}


\section{Introduction}
Machine learning techniques have been widely adopted in the development of autonomous vehicles and advanced driver-assistance systems (ADAS). 
Most autonomous driving and ADAS software stacks, such as Baidu Apollo~\cite{baiduapollo} and OpenPilot~\cite{openpilot}, are generally composed of four-layered modules: sensing, perception, planning, and control~\cite{yurtsever2020survey}. The sensing and perception modules collect data from the surrounding environment via a variety of sensors such as cameras, LiDAR, radar, GPS and IMU, and use learning-based perception algorithms to process the collected data and understand the environment. The planning and control modules leverage the perception results to propose a feasible trajectory and generate detailed commands for the vehicle to track the trajectory. In those systems, deep neural networks (DNNs) are widely used for sensing and perception in transportation scenarios~\cite{gao2018object,xu2020data,siam2017deep}
, such as semantic segmentation, object detection and tracking, as they often provide significantly better average perception accuracy over traditional feature-based methods. For planning and control, there are also increasing interests in applying neural networks with techniques such as reinforcement learning and imitation learning~\cite{rhinehart2018deep}, due to their capabilities of automatically learning a strategy within complex environment. 

However, the adoption of DNN-based techniques in ADAS and autonomous driving also bring significant challenges to vehicle safety and security, given the ubiquitous uncertainties of the dynamic environment, the disturbances from environment interference, transient faults, and malicious attacks, and the lack of methodologies for predicting DNN behavior~\cite{zhu2020know}. In particular, extensive studies have shown that DNN-based perception tasks, such as image classification and object detection, may be susceptible to adversarial attacks~\cite{eykholt2018robust,chen2018shapeshifter}, where small perturbations to sensing input could result in drastically different perception results. There are also recent works on attacking DNN-based perception in ADAS and autonomous vehicles by adding adversarial perturbations to the physical environment in a stealthy way~\cite{zhou2020deep,sato2020hold,song2018physical,eykholt2018robust,sitawarin2018darts}. For instance, \cite{sato2020hold} generates a dirty road patch with carefully-designed adversarial patterns, which can appear as normal dirty patterns for human drivers while leading to significant perception errors and causing vehicles to deviate from their lanes within as short as 1 second. 


The prior works addressing adversarial attacks mostly focus on detecting anomaly in the input  data~\cite{kimin2018simple,zheng2018robust,lu2017safetynet,yin2020adversarial}
or making the perception neural networks themselves more robust against input perturbations~\cite{goodfellow2015explaining,madry2019deep,carlini2017towards}. In ADAS and autonomous driving, however, the impact of adversarial attacks on system safety and performance is eventually reflected through vehicle movement, taking into account of planning and control decisions. Thus, we believe that for those systems, it is important to take a holistic and end-to-end approach that addresses adversarial attacks throughout the sensing, perception, planning and control pipeline.

In our preliminary work recently published in a work-in-progress paper~\cite{liang2021endtoend}, we studied the automated lane centring (ALC) system in OpenPilot~\cite{openpilot}, a popular open-source ADAS implementation, and investigated how the dirty road patch attack from~\cite{sato2020hold} could affect perception, planning and control modules. We discovered that a \emph{confidence score} generated by the perception module could serve as a sensitive signal for detecting such attack, however it does not quantitatively measure the extent of the attack and cannot be effectively used for mitigation. In this paper, motivated by the findings from~\cite{liang2021endtoend}, we propose a novel end-to-end approach for \emph{detecting and mitigating} the adversarial attacks on the ALC system. Our approach quantitatively estimates the \emph{uncertainty} of the perception results, and develops an adaptive planning and control method based on the uncertainty analysis to improve system safety and robustness.  

In the literature, methods have been proposed to address the uncertainties of various modules in the ADAS and autonomous driving pipeline. For instance, the method proposed in~\cite{nakashima2020uncertainty} utilizes estimated uncertainty as a threshold to decide which sensor is reliable. 
In the OpenPilot implementation, Multiple Hypothesis Prediction~\cite{rupprecht2017learning} is utilized to estimate the prediction confidence. Some works propose methods to detect out-of-distribution inputs~\cite{cai2020real} and design probabilistic deep learning based perception models~\cite{sun2018probabilistic} and planning models~\cite{hruschka2019uncertainty}. Different from these prior methods, our approach takes a system-level view and addresses the uncertainty from adversarial attacks throughout sensing, perception, planning and control. While this work focuses on the dirty road patch attack, we believe that our methodology can be applied to other adversarial attacks that cause perception uncertainties, and may be extended to address more general uncertainties (e.g., those caused by environment interference or transient faults).  
Specifically, our work makes the following contributions:
\begin{myitemize}
    \item We analyzed the impact of dirty road patch attack across the ADAS pipeline, and developed a method to quantitatively measure the perception uncertainty under attack, based on the analysis of both model and data uncertainties in the perception neural network. 
    \item We developed an uncertainty-aware adaptive planning and control method to improve system safety and robustness under adversarial attacks. 
    \item We conducted experiments on both public dataset and LGSVL~\cite{rong2020lgsvl}, a production-grade autonomous driving simulator. The results demonstrate that our approach can significantly improve the system robustness over the original OpenPilot implementation when under adversarial attacks, reducing the deviation of lateral deviation by $55\%\sim90\%$.
\end{myitemize}



The rest of the paper is organized as follows. Section~\ref{sec:sys_model} introduces the ALC system in OpenPilot and the adversarial attack model to this system. Section~\ref{sec:our_approach} presents our uncertainty-based mitigation approach to address such adversarial attacks. Section~\ref{sec:experiments} shows the experimental results.

\section{ALC System and Adversarial Attacks}
\label{sec:sys_model}

\begin{figure}[htbp]
\centerline{\includegraphics[width=\columnwidth]{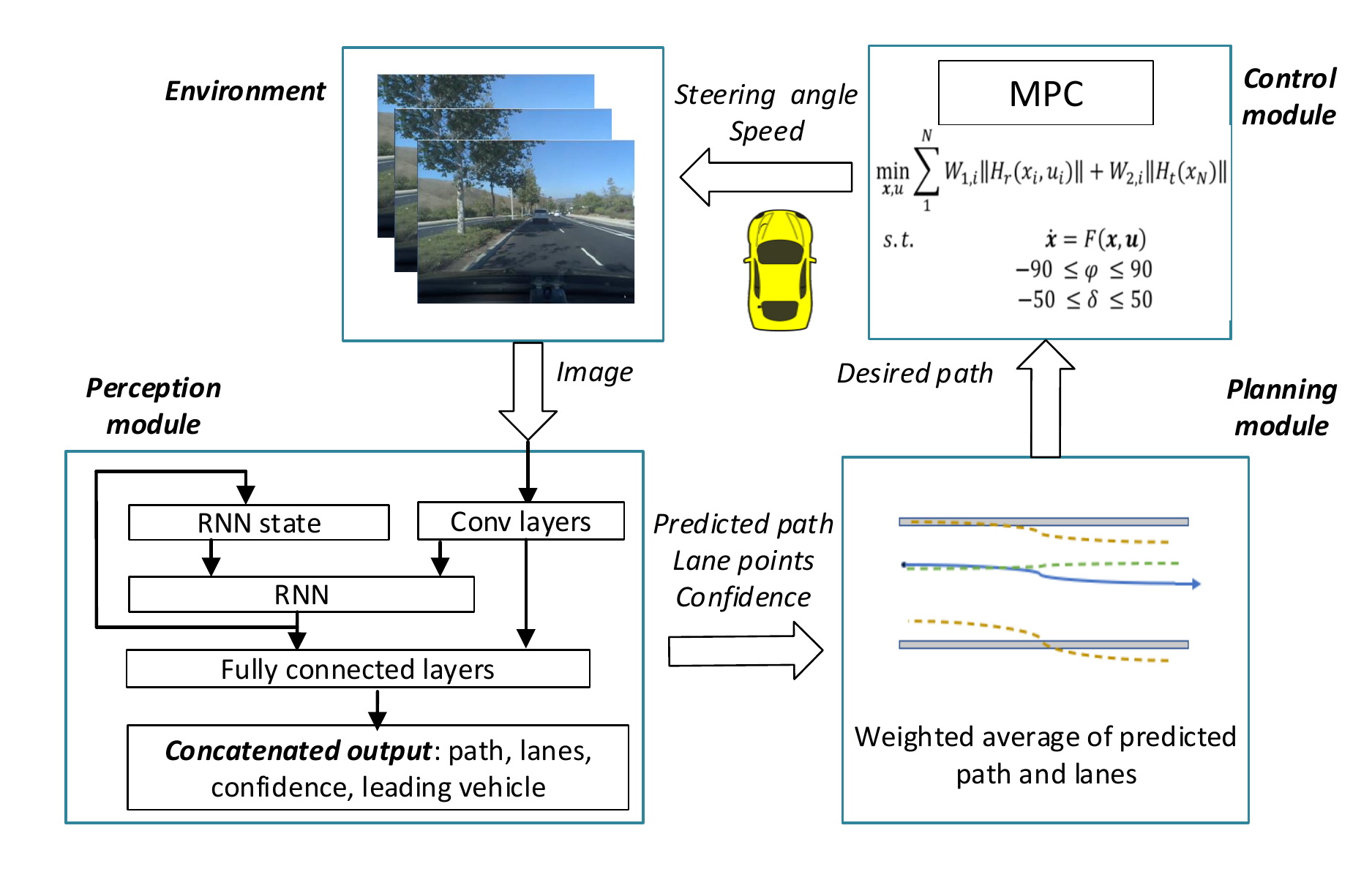}}
\caption{OpenPilot pipeline: the ALC system consists of a DNN-based perception module, a trajectory planning module, and an MPC controller. 
}
\vspace{-12pt}
\label{fig:openpilot}
\end{figure}

The Automated Lane Centering (ALC) system, one of the Level 2 autonomous driving systems, are widely deployed in modern commercial vehicles.  
In the ALC system, the perception module collects vision and distance input from cameras and radars, and outputs the perception of the environment to the planning and control module, which generates a desired trajectory and controls vehicle steering and acceleration.  
In the following, we will take the open-source software Openpilot (Fig.~\ref{fig:openpilot}) as an example to illustrate ALC's architecture.

\subsection{DNN-based Perception Module}
Traditional perception modules often use edge detection or Hough transform to detect lane lines based on camera input~\cite{low2014simple}.
Recently, DNN-based lane detection models achieve the state-of-the-art performance~\cite{wang2018lanenet} and are widely adopted in production-level ALC systems today such as Tesla AutoPilot~\cite{ingle2016tesla} and OpenPilot. The DNN-based perception module can detect lane lines and objects and provide necessary information for the planning and control module. 

In ALC tasks, since the lane lines are continuous across consecutive frames, recurrent neural network (RNN) is typically applied to utilize the time-series information to make the prediction more stable~\cite{zou2019robust}. 
A convolutional neural network (CNN) processes the current frame and  feeds the combination of the CNN's and RNN's outputs into a fully-connect network. The concatenated output includes a predicted path (an estimation of the path to follow based on perception history), the detected right and left lane lines, and corresponding probabilities to indicate the detection confidence of each lane line. Note that the lane detection model first predicts lane points and then fits them into polynomial curves in post-processing for denoise and data compression.

\subsection{Planning and Control Modules}

The planning module generates a trajectory for the vehicle to follow. Sampling based methods, model based methods and deep learning models are often applied to the trajectory planning~\cite{qian2016motion}. For ALC, the desired path generated by the planner should be located in the middle between the left lane line and the right lane line. 
The planner module in OpenPilot generates an estimation of this desired path by calculating a weighted average of the detected left lane line, right lane line and predicted path, with the weights being the confidence scores outputted by the perception module. Intuitively, if the perception module is less confident on the predicted lanes, the generated desired path relies more on the predicted path; otherwise, it will be closer to the weighted average of the predicted left lane line and right lane line~\cite{liang2021endtoend,openpilot}. 

Given the desired path to follow, the lower-level controller calculates the vehicle maneuver and generates commands to control throttle, brake, and steering angle. In OpenPilot ALC, model predictive control (MPC)~\cite{bujarbaruah2018adaptive} is used to calculate the steering angle based on simplified system dynamics, vehicle heading constraints, and maximum steering angle constraints.
MPC-based approach permits high-precision planning and a certain degree of robustness. For longitudinal control, the acceleration is generated by a PID controller by setting an appropriate reference speed.

\subsection{Attack Model 
}
In this paper, we assume a similar attack model as prior work~\cite{sato2020hold}. We focus on  attacks achieved through external physical world. In particular, we assume that the attacker cannot hack through software interfaces nor modify the victim vehicle. However, the attacker can deliberately change the physical environment that is perceived by the on-board sensors of the victim vehicle (e.g., cameras). Furthermore, we assume that the attacker is able to get knowledge of the victim's ADAS/autonomous driving system and can drive the same model vehicle to collect necessary data. This can be achieved by obtaining a victim vehicle model and conduct reverse engineering, as demonstrated in~\cite{tecent2019experimental,checkoway2011comprehensive}.  The attacker's goal is to design the appearance of certain object (in our case, a dirty patch) on the road such that
\begin{enumerate*}
    \item the adversarial object appears as normal/seemingly-benign for human drivers, and
    \item it can cause the ADAS/autonomous driving system to deviate from the driver's intended trajectory.
\end{enumerate*}
Some example attack scenarios are discussed in~\cite{zhou2020deep,sato2020hold}. The work in~\cite{zhou2020deep}  generates an adversarial billboard to cause steering angle error. \cite{sato2020hold} generates a gray scale dirty patch on the road such that vehicles passing through the patch will deviate from its original lane, which is the first attack systematically designed for the ALC system and reaches state-of-art attack effect on production-level ADAS. Through our experiments, we will use the dirty road patch attack~\cite{sato2020hold} as a case study, but we believe that our approach can be extended to other similar physical environment attacks. As discussed in our preliminary work~\cite{liang2021endtoend}, these physical attacks typically will render abnormal behavior in the perception output and then propagate through the entire pipeline.


\section{Our End-to-end Uncertainty-based Mitigation Approach for Adversarial Attacks}
\label{sec:our_approach}

\begin{figure*}[!ht]
    \centering
    \includegraphics[width=1.8\columnwidth]{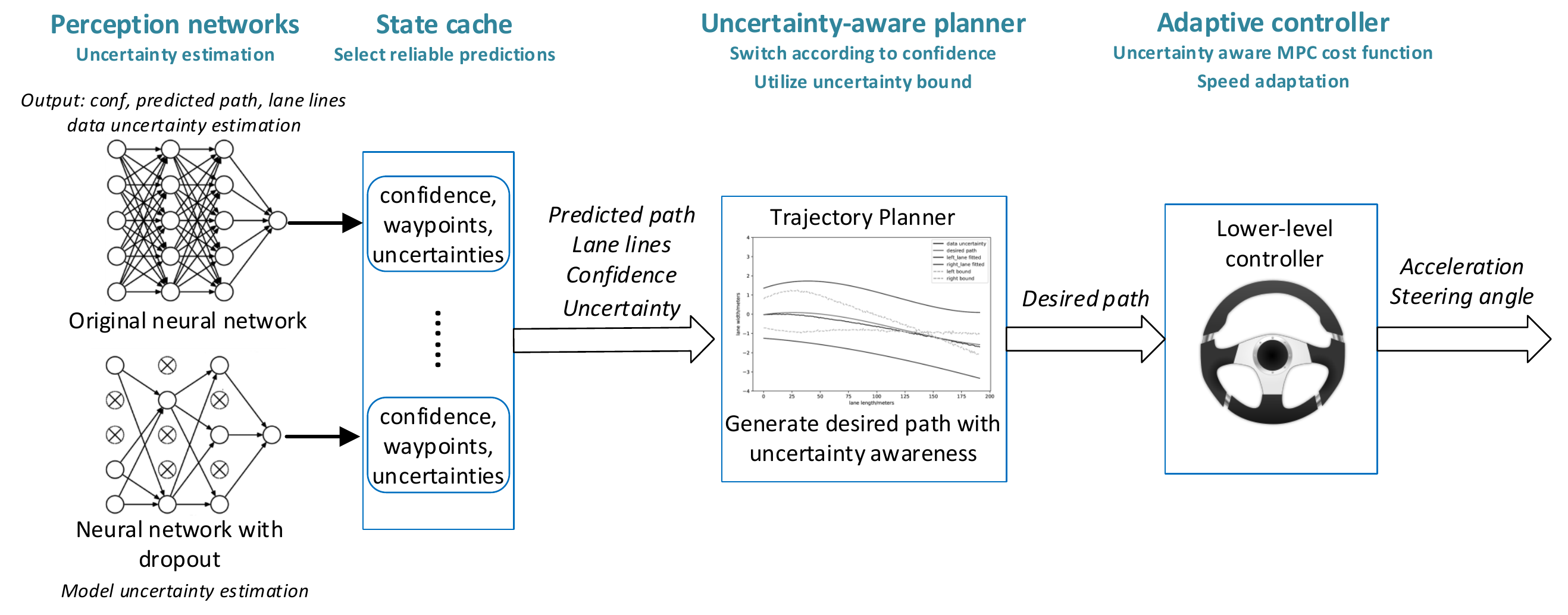}
    \caption{Overview of our proposed ALC pipeline with uncertainty-based mitigation for adversarial attacks.}
    \label{fig:pipeline}
\end{figure*}

As shown later in Fig.~\ref{fig:pipeline}, the perception module in our approach involves two neural network models. The original OpenPilot perception model is used in the normal operation mode (i.e., when the overall prediction confidence is high), where it outputs predicted path, lanes and confidence scores. 
When anomaly is detected, a new perception neural network is used to estimate model uncertainty while generating perception output. 
The trajectory planner will take the uncertainties into consideration and generate a desired path that is less affected by the adversarial attack. Correspondingly, in the lower-level control, more conservative and uncertainty-aware constraints are used in the MPC and a speed adaptation method is applied to ensure safety. The details of our proposed approach are introduced below.


\subsection{Perception Confidence as Signal of Attack}

Measuring the confidence of the DNN's prediction is a significant challenge. In different perception tasks, various methods are applied to estimate the perception confidence. For instance, in YOLO~\cite{redmon2018yolov3}, intersection over union (IOU) measures the confidence of regression. In OpenPilot's neural network, a multiple hypotheses prediction (MHP)~\cite{rupprecht2017learning} classifier is trained with cross entropy loss and its output can represent how confident the lane line is predicted correctly~\cite{ramos2018deconstructing}. 

\begin{figure}
    \centering
    \begin{subfigure}{\columnwidth}
        \includegraphics[width=\textwidth]{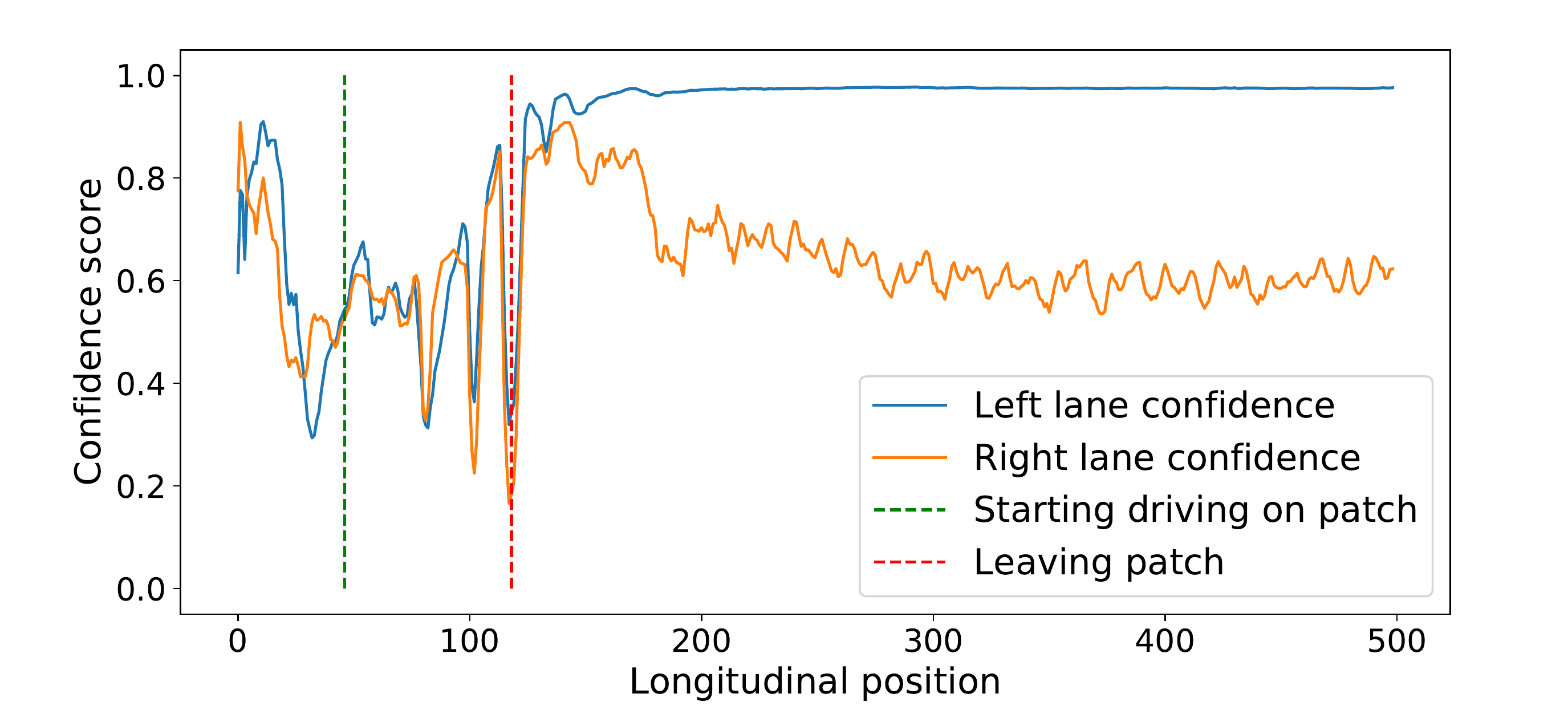}
        \caption{}
        \label{fig:prob_drop}
    \end{subfigure}%
    
    \begin{subfigure}{\columnwidth}
        \includegraphics[width=\textwidth]{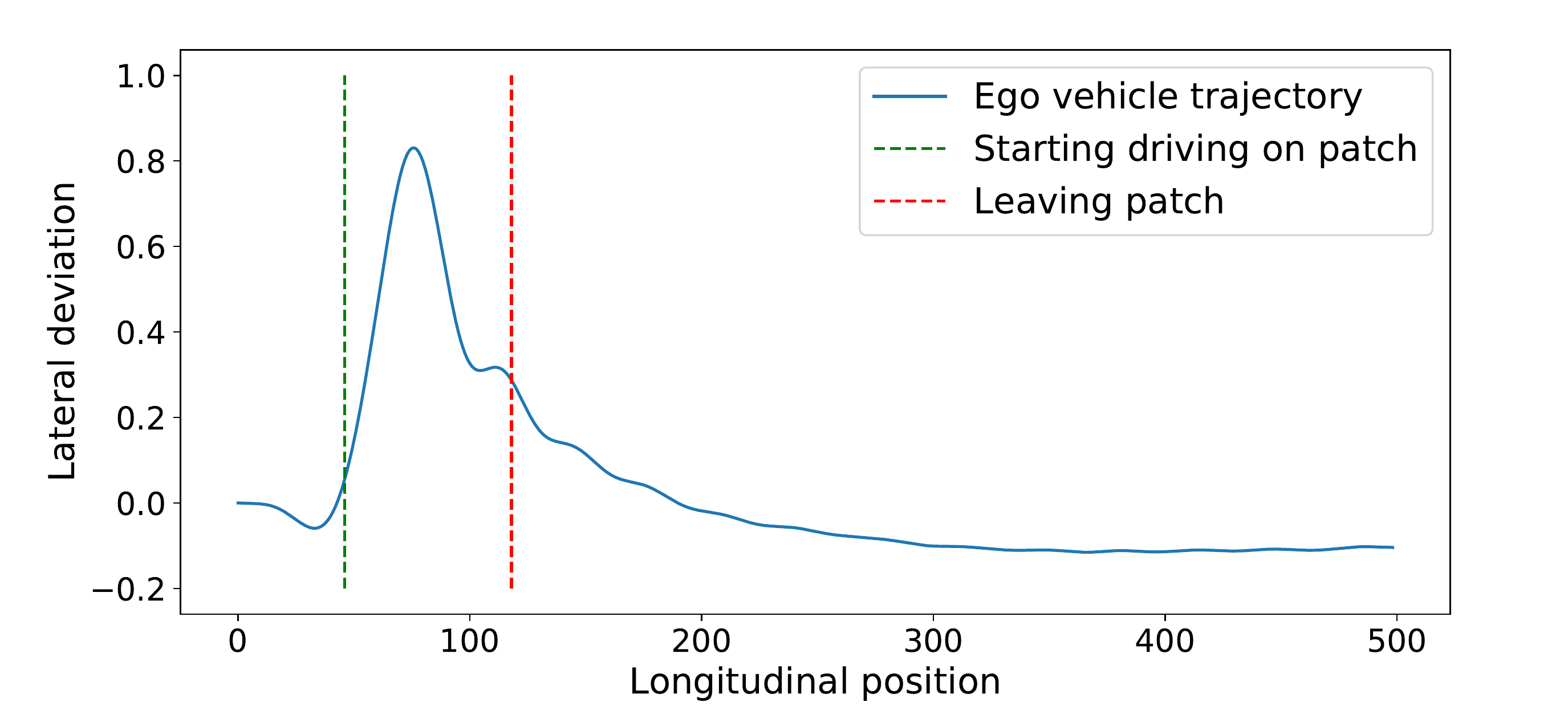}
        \caption{}
        \label{fig:traj_deviate}
    \end{subfigure}
    \caption{The patch starts around 40 meters and is 96 meters long. (a) shows that the confidence of the prediction drops, (b) plots the lateral deviation of the ego vehicle. For more details, please see our preliminary work in~\cite{liang2021endtoend}.
    }
    \label{fig:conf}
    \vspace{-12pt}
\end{figure}

To investigate how the perception confidence affects the system safety, we conduct a series of experiments with different settings of the attacks (early results reported in our preliminary work~\cite{liang2021endtoend}, which focuses on the attack detection alone). As indicated in Fig.~\ref{fig:prob_drop}, we observe a general phenomenon that the confidence score of the perception module drops significantly when the vehicle is approaching the dirty patch, while the confidence score keeps in a relatively stable level in benign cases or under ineffective noises. Fig.~\ref{fig:conf} also shows the consistency between the drop in confidence score and the vehicle's lateral deviation under attack. As discussed in Section~\ref{sec:sys_model}, the desired path generated by OpenPilot's path planner can be considered as a weighted average of the predicted left lane, predicted right lane, and the predicted path, with the confidence scores as weights. Under the impact of the dirty patch, confidence of the predicted two lanes drops, resulting in more weights on predicted path. However, the predicted path deviates from the middle of the line eventually, and the desired path leans towards wrong directions. 

    Based on such observation, we think that the perception confidence can be utilized as a signal to indicate whether the model (perception module) is under adversarial attack. Our mitigation strategy leverages this signal to switch between different perception modules and applies adaptive planning and control accordingly. 
    

\subsection{Uncertainty Estimation and Safety Bound}\label{uncertainty bound}

While the confidence scores generated by the OpenPilot perception module \emph{qualitatively} show the existence of the attack, they do not \emph{quantitatively} measure the extent of the attack and cannot be effectively used for mitigation. Thus, we developed a new approach for quantifying the perception \emph{uncertainty} that considers both data uncertainty and model uncertainty.
Data uncertainty is caused by the input noise/disturbance and intrinsic randomness of the real data while model uncertainty results from a lack of training data in certain areas of the input domain and the test example being out of distribution.  By assuming that the data uncertainty follows a normal distribution, we can estimate it by using maximum likelihood approach. In neural networks, the model can take mean negative log-likelihood function 
as a loss function to capture the aleatory uncertainty~\cite{kendall2017uncertainties}. In Equation~\eqref{log-likelihood} below, $\hat{\mu}(x)$ is the predicted mean value and the distribution variance $\sigma$ is the estimated data uncertainty:

\begin{equation}
\mathcal{L}(x, y)=-\log \phi(y \mid x)=\frac{\log \hat{\sigma}^{2}_{data}(x)}{2}+\frac{(y-\hat{\mu}(x))^{2}}{2 \hat{\sigma}^{2}_{data}(x)}\label{log-likelihood}
\end{equation}

To estimate model uncertainty, similarly to the Bayesian approaches, we consider that there are distributions over the weights of neural networks and the parameters of the distribution are trained on the training dataset. The weight distribution after training can then be written as $p(\omega \mid \mathrm{X}, \mathrm{Y})$. We apply the Monte Carlo methods to estimate the distribution by adding dropout layers to sample weights with dropout rate $\Phi$~\cite{gal2016dropout}:
\begin{equation}
p(\omega | \mathbf{X}, \mathbf{Y}) \approx \operatorname{Bern}(\omega ; \Phi)\label{dropout}
\end{equation}
Therefore, the model uncertainty can be estimated as~\cite{loquercio2020general}:
\begin{equation}
\sigma_{model}^{2} = \frac{1}{T} \sum_{t=1}^{T} \left(\boldsymbol{\mu}_{t}-\overline{\boldsymbol{\mu}}\right)^{2}\label{monte carlo}
\end{equation}
The total variance can be derived by adding the model variance $\sigma_{model}^{2}$ and the data variance $\sigma_{data}^{2}$, as shown in Equation~\eqref{total error}.
To the best of our knowledge, current commercial driving assistance systems do not take the perception uncertainty into consideration. 
\begin{equation}
\sigma_{total}^{2} = \sigma_{model}^{2}+\sigma_{data}^{2}\label{total error}
\end{equation}

\begin{figure}[!ht]
    \centering
    \includegraphics[width=\columnwidth]{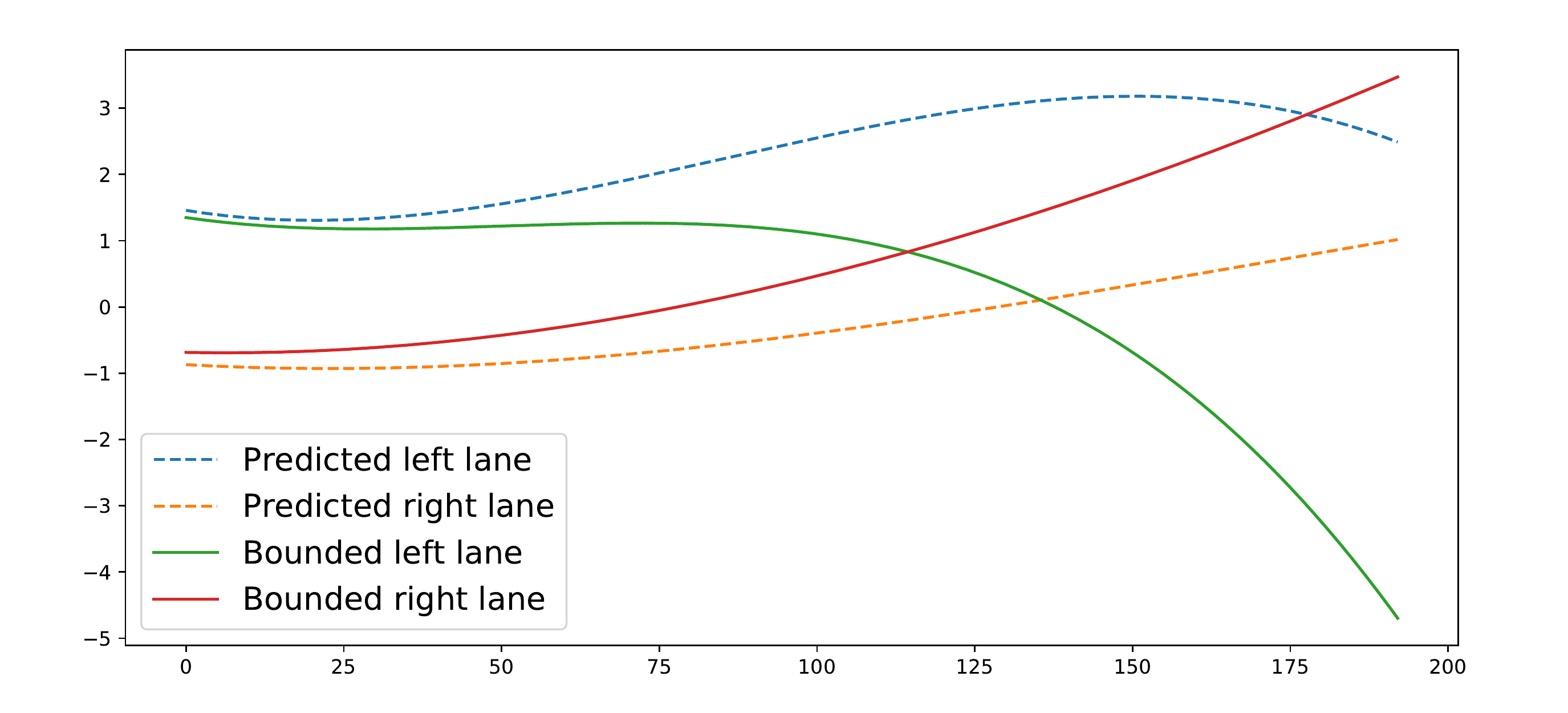}
    \caption{The original OpenPilot prediction and the bounded line lanes after compensating the original prediction with estimated uncertainties. 
    }
    \label{fig:demo_bound_predict}
\end{figure}

When the overall prediction confidence drops below a threshold, we deem that the ego vehicle is experiencing significant environment noise or under adversarial attack. Instead of using the original prediction result, we leverage the obtained prediction uncertainties to more conservatively measure the bounds of left lane and right lane. Let $\mu_{i}$ be the predicted mean of the $i$-th point of the lane and $\sigma_{total,i}$ be the corresponding error, then $[\mu_{i}-\sigma_{total,i},\mu_{i}+\sigma_{total,i}]$ can be considered as the range where the ``true'' lane point may reside. Specifically, we use $\mu_{i}-\sigma_{total,i}$ as the lower bound of left lane if $\mu_{i}$ is a point on the left lane while $\mu_{i}+\sigma_{total,i}$ as the upper bound if it is a point on the right lane. We use a polynomial function $p(i)$ to approximate the points along the lane. The problem can be cast as a weighted least square fitting. For left lane, we fit the curve by minimizing $\sum_{i}^{N}\omega_{i}|p(i)-(\mu_{i}-\sigma_{total,i})|$; for right lane, we minimize $\sum_{i}^{N}\omega_{i}|p(i)-(\mu_{i}+\sigma_{total,i})|$. Here, $\omega_{i}$ is the weight. Since points further away tend to exhibit larger uncertainties, we assign closer points with larger weights by setting $\omega_{i} = \frac{1}{\sigma_{data}}$. Fig.~\ref{fig:demo_bound_predict} shows the bounded predicted lanes using our approach. As we can see, the bounded prediction is more conservative compared to the original approach. In most cases, the bounded left lane and right lane tend to intersect at some point. 

\subsection{Uncertainty-aware Trajectory Planner}
In the original design of OpenPilot, the desired path can be considered as a weighted sum of the left lane, right lane and predicted path (generated by the perception module). Lanes/path with higher confidence score will be assigned with larger weights and have more impact on the desired path. However, we find out that this approach cannot handle adversarial scenarios. Considering that the attacker can manipulate sensor data such that the confidence scores of the predicted left lane and right lane are much lower while the predicted path is bent towards left. In this case, the desired path will also lean towards left even though the vehicle is driving on a straight road. We argue that, when the confidence score drops below certain threshold, we should explicitly consider the uncertainties instead of just relying on the predicted road curves. That is, we use the uncertainty-bound area derived above to constrain vehicle's speed and steering angle.

The pseudo-code to calculate the desire path in our uncertainty-aware approach is shown in  Algorithm~\ref{alg:cal_d_poly}. Lines 5-6 calculate the accumulated uncertainties along the left lane and right lane. $\omega_{left}$ and $\omega_{right}$ are the corresponding weights assigned to the bounded left lane and bounded right lane. $p_{weighted}$ is the weighted path and the final desired path is the weighted average of $p_{weighted}$ and $p_{openpilot}$ (the latter is the desired path obtained by running the original OpenPilot). The intuition is that if the current prediction has low overall confidence, we will rely more on the uncertainty-aware bounded prediction to mitigate the adversarial attack.
\begin{algorithm}[!h]
\caption{DesiredPath: Uncertainty-aware Desired Path Calculation}
\label{alg:cal_d_poly}
\begin{algorithmic}[1]
\REQUIRE \text{Polynomial for lane lines:
$p_{l}, p_{r}$;}
\text{the overall confidence of the prediction: $lr_{conf}$}
\IF{total confidence $lr_{conf}$ is less than 
$Conf\_Threshold$}
    \STATE use the original Openpilot to generate the desired path $p_{openpilot}$
    \RETURN $p_{openpilot}$
\ELSE
    \STATE $left_{sum} = \sum_{i=1}^{n}\sigma_{left,i}$ 
    \STATE $right_{sum} = \sum_{i=1}^{n}\sigma_{right,i}$ 
    \STATE $total_{sum} = left_{sum}+right_{sum}$
    \STATE $\omega_{left} = \frac{right_{sum}}{total_{sum}}$
    \STATE $\omega_{right} = \frac{left_{sum}}{total_{sum}}$
    \STATE $p_{weighted} = \omega_{left}*p_{l}+\omega_{right}*p_{r}$
    \STATE $desired\ path = (1-lr_{conf})*p_{weighted}+lr_{conf}*p_{openpilot}$ 
\ENDIF
\RETURN $desired\ path$
\end{algorithmic}
\end{algorithm}

Besides utilizing the estimated uncertainty to bound the safe trajectory area and produce the desired path, we also make use of the temporal locality of the path prediction. We notice that the perception and planning modules can produce a safe desired path for about 100 meters with frequency of 20 Hz while the vehicle will only move forward up to several meters in the period. Therefore, the information of consecutive frames has considerable locality and relevance. We maintain a state cache to store the perception output of most recent $k$ consecutive frames. In our experiment, we pick $k=7$ to store the information of the past $0.35$ seconds. In case of adversarial scenarios, the system will select the perception output with highest confidence score from the state cache as the planner input. Taking advantage of the locality, the system robustness to short-term inference will be improved.  

\subsection{Adaptive Controller}
\subsubsection{Uncertainty-aware cost function}
An MPC controller is used to generate an appropriate steering command based on the ego vehicle status and the desired path. We modify the MPC controller by explicitly considering the more conservative uncertainty-aware bounded lanes. The optimization objective can be written in the form of the summation of a running cost and a terminal cost:
\begin{equation}
    \min_{\bold{x},\bold{u}}\quad \sum_{i = 1}^{N} \bold{W}_{1,i}\| \bold{H}_{r}(x_{i},u_{i})\|^{2} + \bold{W}_{2,i}\|\bold{H}_{t}(x_{N})\|^{2}
\end{equation}
where $\bold{W}_{1,i}$ and $\bold{W}_{2,i}$ are weight matrices; $\bold{H}_{r}$ is the reference function to capture the difference between current ego vehicle states and the desired path. $\bold{H}_{t}$ is a measurement function regarding the ego vehicle states at the end of prediction horizon. Intuitively, given the desired path, we want the vehicle to drive along the reference path, but we also want the vehicle to driven on the center of traffic lanes. This is achieved by considering the the distance errors with respect to the  desired lane and traffic lanes. However, we argue that the distance error regarding the left lane and right lane should adopt the bounded prediction instead of the original OpenPilot prediction. Specifically, the reference function is written as:
\begin{align*}
    &\bold{H}_{r}(x_{i},u_{i}) = \\
    &\begin{bmatrix}
    p_{d}(x_{i})-y_{i}&
    e^{-(p_{l}(x_{i})-y_{i})}&
    e^{p_{r}(x_{i})-y_{i}}&
    \epsilon_{h}&
    \epsilon_{a}&
    u^{2}
    \end{bmatrix}^{T}
\end{align*}
where $p_{d}$ is the polynomial representing the desired path, $p_{l}$ and $p_{r}$ are fitted polynomial representing bounded left and bounded right lane respectively. $\epsilon_{h}$ and $\epsilon_{a}$ are the error regarding the ego vehicle heading and angular rate respectively. $u^{2}$ is a penalty term to avoid aggressive steering. The measurement function $\bold{H}_{t}$ is similar as $\bold{H}_{r}$, except that it does not consider angular rate and penalty. The constraints include
\begin{enumerate*}
\item system dynamics
\item vehicle heading is limited to $[-90^{\circ},90^{\circ}]$ and
\item the maximum steering angle is $50^{\circ}$.
\end{enumerate*}
\subsubsection{Speed Adaptation}
Generally, the uncertainty for further-way points tends to be considerably large (e.g., Fig.~\ref{fig:uncertainty estimation}) even for benign driving scenario. This means that, for safety consideration, the ego vehicle is not sure about the road structures that are far away. To prevent the vehicle from driving too fast under uncertain scenarios, we apply an emergency brake to the vehicle if its current speed is too fast. Specifically, when the overall prediction confidence drops below the threshold $Conf\_Threshold$, a maximum deceleration $\alpha_{max}$ is applied. The entire end-to-end pipeline of our approach is shown in Algorithm~\ref{alg:pipeline}. Lines 1-3 calculate prediction uncertainties. Then the result is pushed into state cache. If the confidence of the prediction is too low, we pick the perception result with the highest confidence score from the state cache (line 5-7). Then the bounded predicted lane curves are used to calculate the desired path as in Algorithm~\ref{alg:cal_d_poly}. Moreover, if the confidence score drops below the threshold, we apply an emergency brake to slow down the vehicle. Here, $v_{min}$ is the minimum speed required. Finally, the MPC controller calculates the steering angle for the next period (line 12).
\begin{algorithm}[htbp]
\caption{Our Uncertainty-aware pipeline for ALC}
\label{alg:pipeline}
\SetAlgoNoLine
\begin{algorithmic}[1]
\REQUIRE 
\text{Current speed $v_{current}$, reference speed $v_{ref}$}
\text{and input image $Input$}
\STATE $conf, {pts}_{lane}, \sigma_{data}^{2} =\boldsymbol{NN}(Input)$
\STATE $\sigma_{model}^{2} = \boldsymbol{Monte\_Carlo\_Dropout}(Input)$
\STATE $\sigma_{total}^{2} =  \sigma_{model}^{2}+\sigma_{data}^{2}$
\STATE push $conf, {pts}_{lane}$ into $state\_cache$ 
\IF{$conf < Conf\_Threshold$}
    \STATE $pts_{lane} = \mathop{\arg\max}_{conf}( state\_queue)$
    \ENDIF
\STATE $desired\_path = \boldsymbol {DesirePath}(pts_{lane},conf,\sigma_{total}^{2})$

\IF{$conf < Conf\_Threshold$ }
    \STATE ${v}_{ref} = min(v_{current}-\alpha_{max}*\Delta t, {v}_{min})$
    \ENDIF

\STATE $steering\_angle, acc = \boldsymbol{MPC}(desired\_path, v_{ref})$
\RETURN $steering\_angle, acc$
\end{algorithmic}
\end{algorithm}
\section{Experimental Results}
\label{sec:experiments}

We implement our design within the open-source driving assistance system OpenPilot. We test our proposed approach using both open dataset (e.g., comma2k19 dataset~\cite{schafer2018commute}) and
synthetic scenarios with an autonomous driving simulator.

For the open dataset, we adopt the kinematic bicycle model~\cite{kong2015kinematic} as the motion model for the ego vehicle. The kinematic bicycle model is commonly used to simulate how vehicles move according to speed and the front steering angle. The kinematic model can produce the vehicle trajectory and obtain the lateral deviation to measure the effectiveness of our proposed mitigation strategy. Moreover, as the trajectory calculated by the motion model may deviate from the original trace, we adopt the motion model based input generation from~\cite{sato2020hold} that combines motion model with perspective transformation to dynamically calculate camera frame updates due to attack-influenced vehicle control. 


For synthetic scenarios, we combine OpenPilot with LGSVL~\cite{rong2020lgsvl} to set up a closed-loop simulation environment. LGSVL is a Unity-based multi-robot simulator developed by LG Electronics America R\&D Center.  We reuse the LGSVL-OpenPilot bridge~\cite{sato2020hold} to transfer sensor data and driving command between the LGSVL simulator and OpenPilot. For both simulation methods, we test the original OpenPilot and ofur proposed design under different attacks as well as in benign situation.

\subsection{Adversarial Attacks}
\begin{figure}[htbp]
\centering
\begin{minipage}[t]{0.4\columnwidth}

\includegraphics[width=1\linewidth]{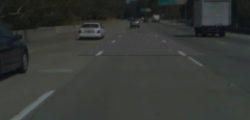}

\end{minipage}
\begin{minipage}[t]{0.4\columnwidth}
\centering
\includegraphics[width=1\linewidth]{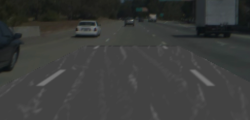}

\end{minipage}
\caption{Input images in benign scenario and under attack. 
}
\label{fig:input}
\end{figure}

In our work, as mentioned in previous sections, we applied the optimization-based physical attack in \cite{sato2020hold}. 
The attack works by placing an optimized patch on the road and it can lead the vehicle out of the lane within 1 second (which means the lateral deviation is larger than 0.735m in highway). The attack against OpenPilot shows high success rate and significant attack effect. To analyze the safety of the system and evaluate our proposed design, we conduct experiments under different settings (perturbation area, stealthiness levels, etc). The benign and attacked inputs are shown in Fig.~\ref{fig:input}.

\subsection{Results for Uncertainty Estimation}
\begin{figure}[htbp]
\centering
\begin{minipage}[b]{0.49\columnwidth}

\includegraphics[width=1\linewidth]{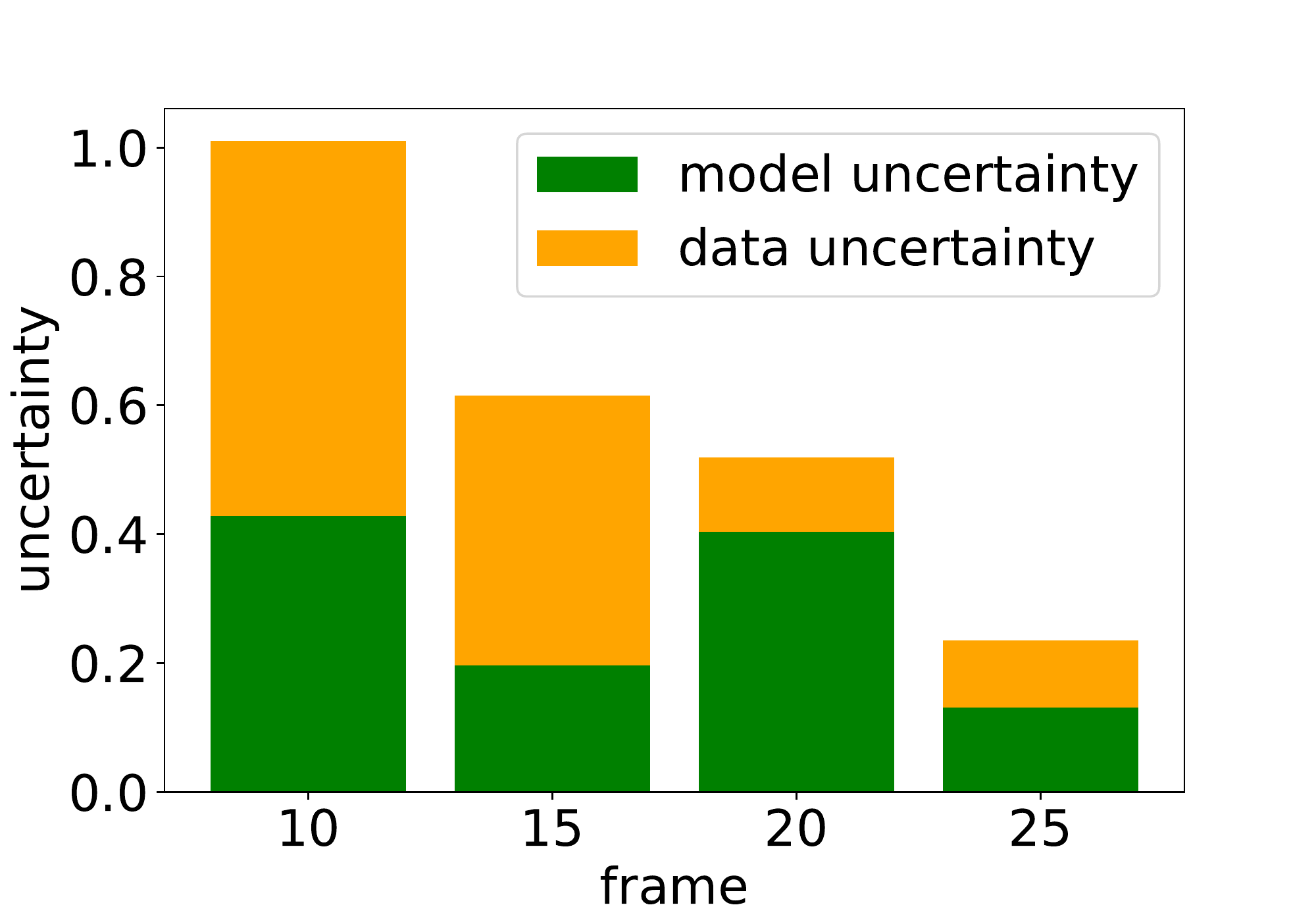}

\end{minipage}
\begin{minipage}[b]{0.49\columnwidth}
\centering
\includegraphics[width=1\linewidth]{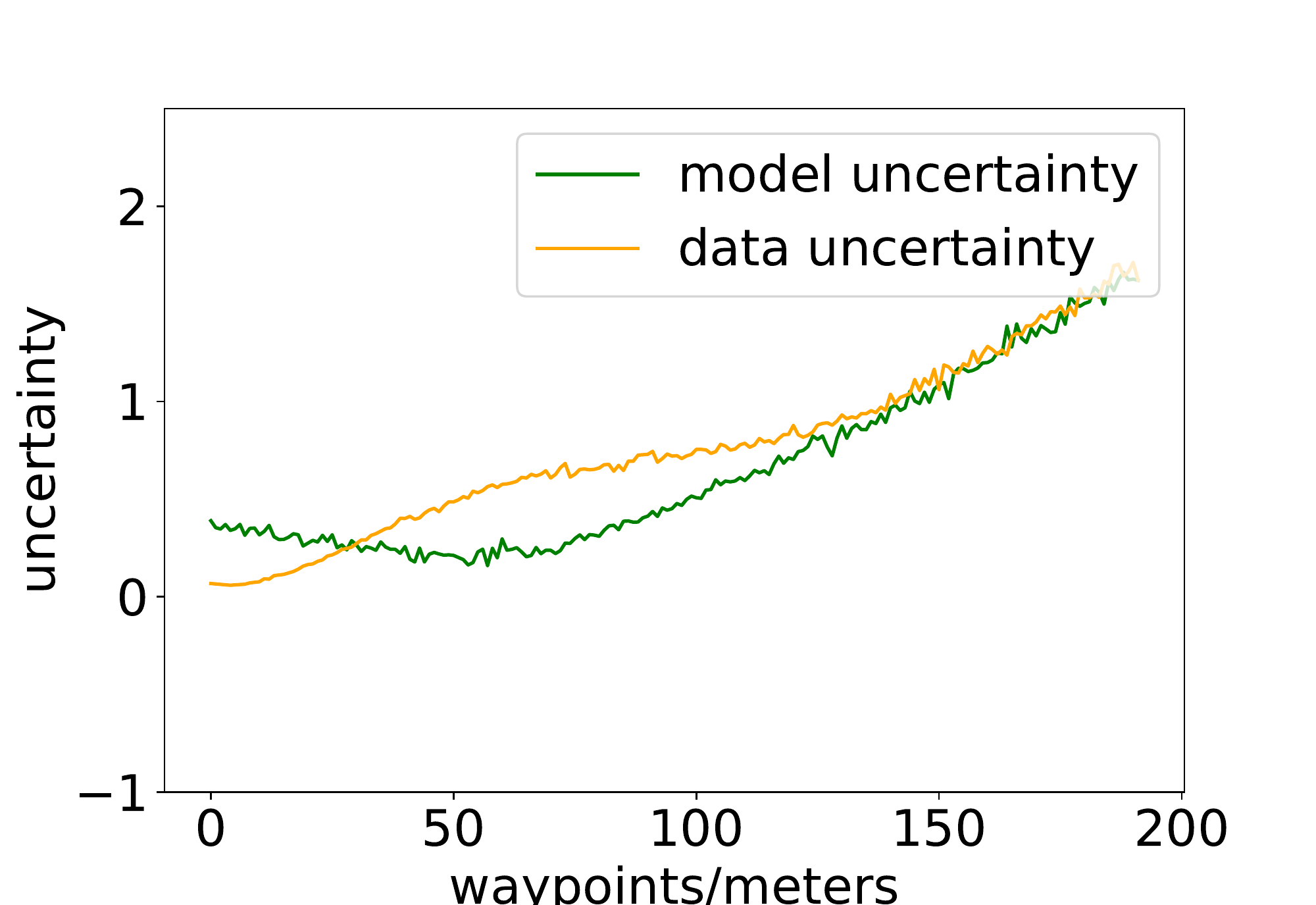}

\end{minipage}
\caption{Data uncertainty and model uncertainty. The left figure shows the overall uncertainties in different frames. The right figure shows the uncertainties increase with the distance. 
}
\label{fig:uncertainty estimation}
\end{figure}

\begin{figure}[!ht]
    \centering
    \includegraphics[width=\columnwidth]{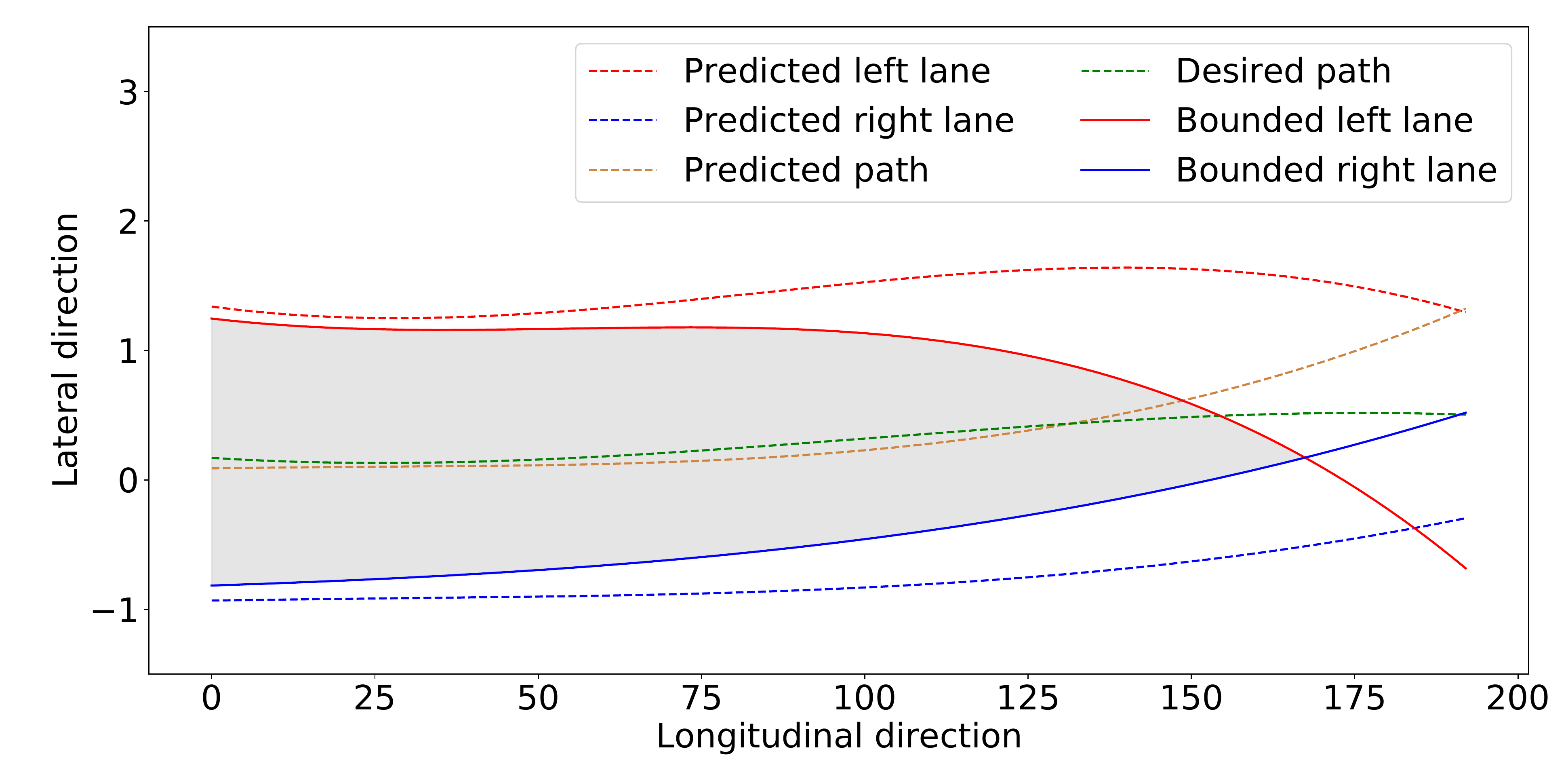}
    \caption{The shaded gray area indicates the safe area bounded by our estimation of uncertainties. The bounded area doesn't deviate to left under attacks.  
    }
    \label{fig:demo_bound_predict_desire}
\end{figure}

For each input frame, our system will calculate the data uncertainty by distributional parameter estimation and model uncertainty by Monte Carlo Dropout methods. The distributional parameter estimation is embedded in the original OpenPilot perception neural network. For the Monte Carlo Dropout, we only activate dropout in inference. There is always a trade-off between inference speed and estimation accuracy: larger number of samples results in higher accuracy and longer time. According to the analysis in~\cite{loquercio2020general} and our experiments, we set the dropout rate as 0.2 and the number of samples as 20. In this setting, the system can obtain precise variance estimation and reach a processing frequency of about 20Hz. The system will estimate uncertainties for 192 points of the predicted lane line in every frame. In experiments, we find that in most cases, data uncertainty and model uncertainty are at roughly the same order of magnitude (left subgraph in Fig.~\ref{fig:uncertainty estimation}). The uncertainty is relatively large when the vehicle is approaching the adversarial patch, which is consistent with the observations in confidence estimation. An example for our uncertainty estimation is shown in Fig.~\ref{fig:uncertainty estimation}. In a single frame, the uncertainties increase with the distance from the current position. This observation facilitates the planner to estimate a safe bound. As shown in Fig.~\ref{fig:demo_bound_predict_desire}, the uncertainty bound will form a triangle-like safe area and we will get a desired path from our planner, which is less affected by the attacks.

\subsection{Mitigation Results for Open Dataset}
\begin{figure}[htbp]
\centerline{\includegraphics[width=\columnwidth]{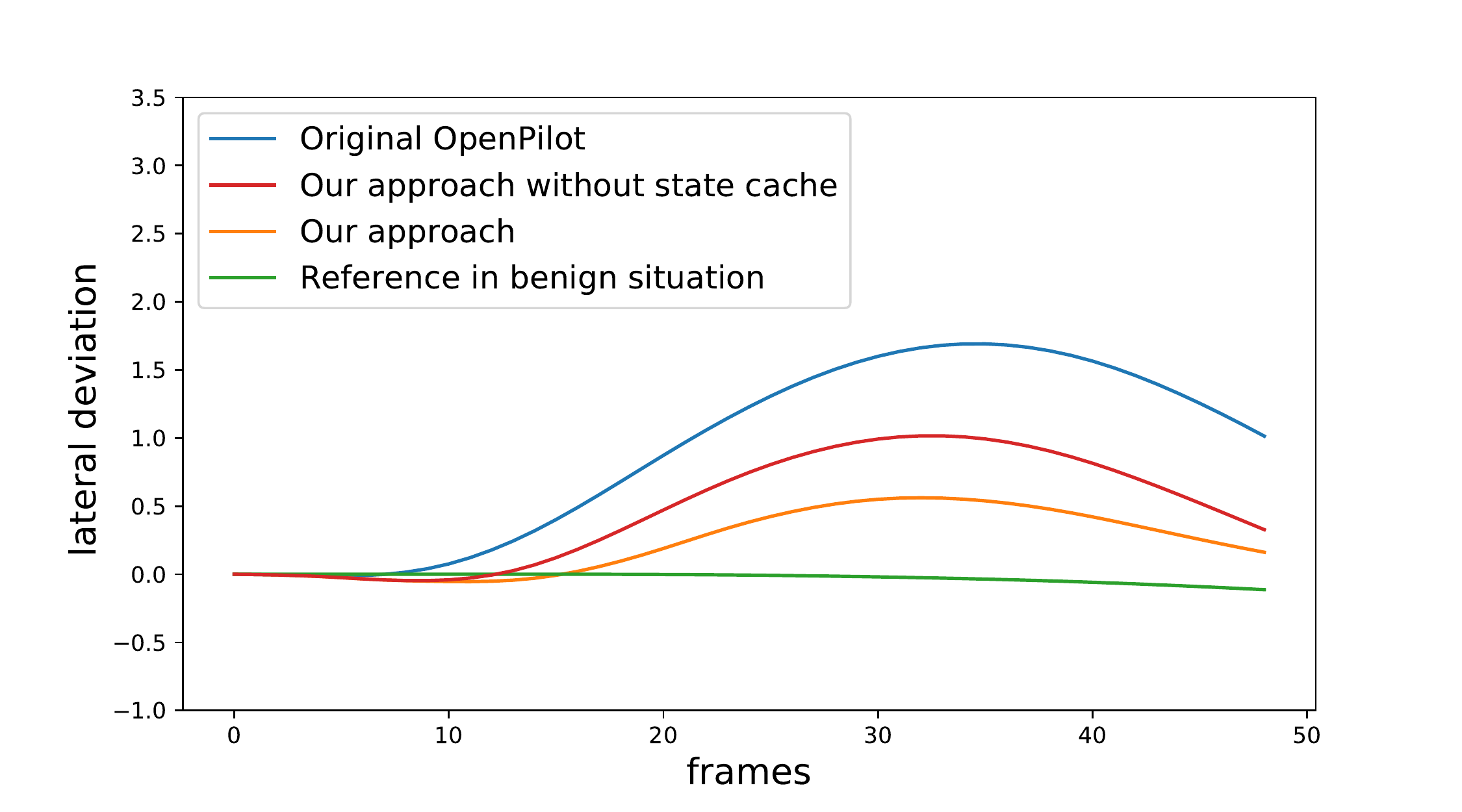}}
\caption{Comparison of our proposed method (with or without state cache) with the original OpenPilot and reference benign situation in terms of lateral deviation. The y-axis positive direction means left. 
}
\label{fig:lateral dev}
\end{figure}

We first test our proposed pipeline with highway image dataset and kinematic model. The input is consecutive image frames captured in a straight four-lane highway and the vehicle's speed is 20 m/s. In the benign case, our proposed systems and the original OpenPilot can both drive in the center of the lane. Fig.~\ref{fig:lateral dev} shows that the adversarial attacks can lead the vehicle with the original OpenPilot out of road and the deviation is more than 1.5 meters. In contrast, the system with our proposed uncertainty estimation and adaptive uncertainty-aware planner and controller can significantly mitigate the attack's effect, reducing the lateral deviation by about $66.8\%$. Besides, the experiments specifically demonstrate that the adversarial attack's effect can be further mitigated by utilizing the temporal locality with state cache. However, in this simulation setting, it is difficult for us to change the vehicle's speed and perception sampling frequency since they were determined by the driving scenario and sensor configurations in the dataset, which motivates us to conduct the following closed-loop simulation in LGSVL.

\subsection{Mitigation Results in LGSVL-OpenPilot Environment}
\begin{figure}[!htbp]
    \centering
    \includegraphics[width=\columnwidth]{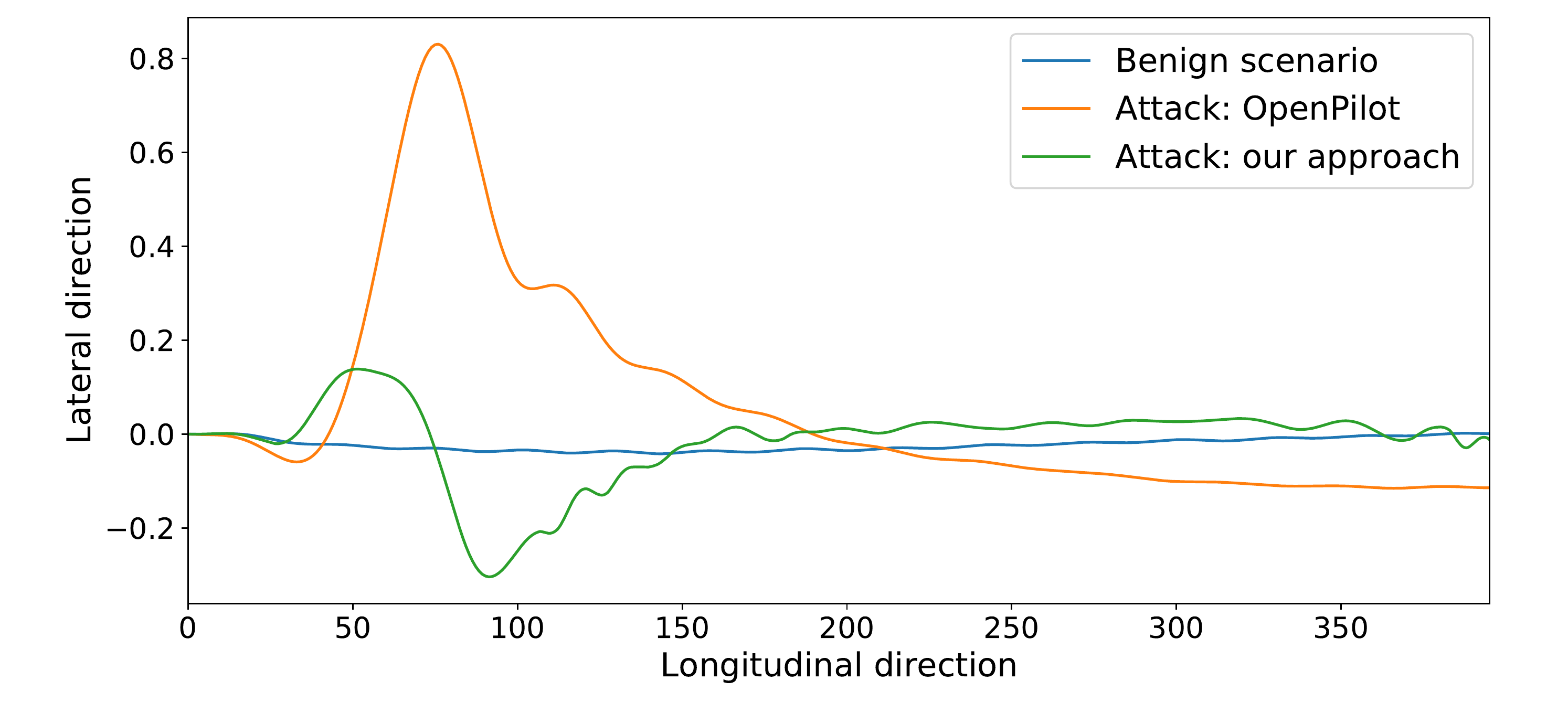}
    \caption{Ego vehicle trajectory under different scenarios. 
    }
    \label{fig:lgsvl_speed_adapt}
\end{figure}

In the closed-loop LGSVL simulation, we set the vehicle to follow a reference speed (20m/s) and control the throttle using a PID controller.  Fig.~\ref{fig:lgsvl_speed_adapt} compares our proposed approach with the original OpenPilot under adversarial scenario. The baseline is the solid blue curve which is the trajectory of the original OpenPilot driving under benign scenario. The solid green curve is the trajectory of the original OpenPilot driving under adversarial scenario where the patch is placed at 40 meters. As we can see, the original OpenPilot does not take prediction uncertainties under consideration and can deviate significantly under adversarial attack. Our proposed is also affected by the patch on the road but can significantly alleviate the adversarial impact.

To further evaluate our proposed approach, we conduct the experiments with different settings of the patch. We place the patch on different positions along the road (40m, 80m and 120m along the longitudinal direction) and adapt the perturbation area ratio (PAR) of the patch. The perturbation area ratio denotes the percentage of the pixels on the patch that are allowed to be perturbed. For our experiment, we generate dirty road patch with PAR ranging from $25\%$ to $100\%$. The maximum lateral deviation varies depending on different experiment settings. Throughout our experiments, the lateral deviation of the original OpenPilot ALC ranges from $0.8$ to $1.2$ meters, which is large enough to drive the victim vehicle out of the lane boundary. In contrast, the lateral deviation of our proposed approach ranges from $0.1$ to $0.4$ meter (still within the lane boundary). Overall, our approach can reduce the lateral deviation by $55.34\%\sim 90.76\%$, under various patch settings.

\section{Conclusions}

In this work, we proposed a novel end-to-end uncertainty-based mitigation approach for adversarial attacks to the automated lane centering system. Our approach includes an uncertainty estimation method considering both data and model uncertainties, an uncertainty-aware trajectory planner, and an uncertainty-aware adaptive controller. Experiments on public dataset ad a production-grade simulator demonstrate the effectiveness of our approach in mitigating the attack effect. We believe that our methodology can be applied to other ADAS and autonomous driving functions, and will explore them in future work.



\bibliographystyle{IEEEtran}
\bibliography{reference}

\end{document}